\documentclass[11pt]{article}
\usepackage[letterpaper,margin=1in]{geometry}
\usepackage{times}
\usepackage{amsmath,amssymb,booktabs,float,graphicx,microtype}
\usepackage[round]{natbib}
\usepackage{hyperref}
\usepackage{url}
\hypersetup{hidelinks}

\title{Share the Judge, Learn the Deferral:\\
Where Specialization Helps LLM Evaluation}
\author{Ye Chen\\
Alibaba Group\\
\and
Weining Zhang\thanks{Corresponding author.}\\
Cheung Kong Graduate School of Business\\
}
\date{}

\begin{document}
\maketitle

\begin{abstract}
Agentic systems produce candidate outputs faster than anyone can review
them. We ask where domain specialization belongs in an LLM evaluator: in
the judge's weights, or in the rule that decides when its judgment can be
accepted. On 99,952 public rubric-conditioned examples, the correct rubric
raises accuracy on a frozen test split by 2.11 points over a rubric-ablated
control. An unrelated rubric costs 2.66 points. Splitting the same training
data across eight criterion-family LoRA experts loses 10.05 points, and
risk-controlled coverage at a 5\% error bound falls from 24.44\% to 5.43\%.
Stored capacity, learning rate, and step count do not explain the loss;
starting the experts from a shared trained judge recovers 19.94 points of
it. A budget sweep from 372 to 5,955 examples per family finds no point
where scratch specialization pays, at either of two granularities. A
warm-started split looks free next to the same-budget monolith, but a
matched control shows the real price: below roughly 3,000 examples per
family, one more pooled epoch wins, and the split first pulls ahead where
pooled training stops improving. The mechanism also holds under real
expert labels: on HealthBench, where physicians wrote the criteria and
judged the responses, the rubric adds 1.60 points and a wrong rubric costs
4.31.

Specializing the deferral decision behaves differently. On RewardBench 2,
small logistic deferral heads route examples through a 0.6B--4B--8B
reward-model cascade without touching any score. Across 20 frozen
repartitions the cascade reaches 89.40\% accuracy, against 84.75\% for the
8B judge alone, at 0.415 of its compute. Every run passes an exact 95\%
risk audit. Margin-based deferral stays near 84.8\% and needs at least 0.94
compute. The gain crosses model families: the same acceptance decisions
land 21.9 points above a weaker Tulu-3-8B terminal built on a different
recipe, and a 184M DeBERTa first stage lifts the Skywork-8B terminal to
87.95\% at 0.723 compute. The design rule is conditional. Pool judgment
training, or split only from a shared start. Specialize the deferral
decision instead: a small head learned from each stage's own scores, under
an audited acceptance bound, as long as the earlier judges are cheap.
\end{abstract}

\section{Introduction}

Language-model agents can produce drafts, plans, or tool trajectories much
faster than a knowledgeable reviewer can read them. Demonstrations hide the
imbalance because generation is the visible event; review is a line in the
pipeline. In deployment the answer must also satisfy a local policy, a
professional standard, or a task-specific rubric, and the review step becomes
the limiting resource.

A common response is to put a larger model behind the agent as a judge.
That improves judgment without settling the operational question. The judge
is called at roughly the rate of generation. A scalar score does not record
which rule governed the decision. Aggregate accuracy says nothing about which
individual cases are safe to accept without further review. An evaluator that
knows when to abstain is often worth more than a slightly more accurate one
that does not \citep{geifman2017selective}.

The domain knowledge involved is usually already written down. Firms keep
rubrics, review checklists, operating procedures, and corrected decisions,
and the standard practice is to paste them into the prompt of a general
judge. We test a different use: compile the recurring criteria into a small
evaluator built from a shared base model, criterion-family adapters, a router
derived from rubric text, and separately audited acceptance thresholds.

We call the adapters \emph{judgelets}: LoRA experts in a routed adapter bank,
each trained on a family of related criteria rather than on a separate
foundation model \citep{hu2022lora}. Only the routed adapter is active for a
given example, so stored specialization can grow without inflating inference
cost. The design also makes three decisions explicit: which rule applies,
which adapter evaluates it, and whether the resulting score is reliable
enough to accept.

Judgelets are an experiment, not a recommendation. The first study concerns
direct rubric-conditioned scoring. We partition the Prometheus Feedback
Collection into training, calibration, development, frozen-test, and
held-criterion sets, then compare rubric-ablated, monolithic, and
sparse-adapter evaluators. A rubric-replacement intervention checks whether
an evaluator actually reads the criterion text. The second study concerns
deferral. Four model sizes from one open reward-model family let us ask
whether a small learned correctness head can cut evaluator compute on
RewardBench 2 without giving up accuracy.

Several results run against the intuition that motivated the design. The
rubric-ablated Prometheus control is strong. A raw-margin cascade misses its
risk target. The eight-family judgelet bank is much worse than one monolithic
adapter. We keep the negative result and use it to locate a boundary: small
evaluators need shared judgment training, while specialized deferral can
exploit differences between judges that are already trained.

Three follow-up studies probe the obvious objections. A data-budget sweep
quantifies how much data a family split needs before it stops hurting.
Cross-family cascades test whether the deferral result depends on all stages
sharing one training recipe. A physician-labeled slice of HealthBench tests
the rubric mechanism under real expert labels rather than synthetic ones.

The paper makes four empirical contributions:
\begin{enumerate}
    \item a controlled rubric study with a frozen unseen-criterion split, a
    rubric-ablated control, and a counterfactual rubric intervention,
    replicated on physician-labeled HealthBench data;
    \item a granularity study that separates sparse deployment from the
    statistical cost of fragmenting supervision, with capacity-, step-, and
    learning-rate-matched controls and a shared-initialization remedy;
    \item a granularity-by-budget grid with a matched continued-training
    control that turns ``enough data to justify a split'' into numbers; and
    \item a deferral study with exhaustive stage subsets, four confidence
    baselines, disjoint fitting, calibration, and test partitions, exact
    held-out risk audits, and cross-family transfer tests.
\end{enumerate}

\begin{figure}[t]
\centering
\includegraphics[width=\linewidth]{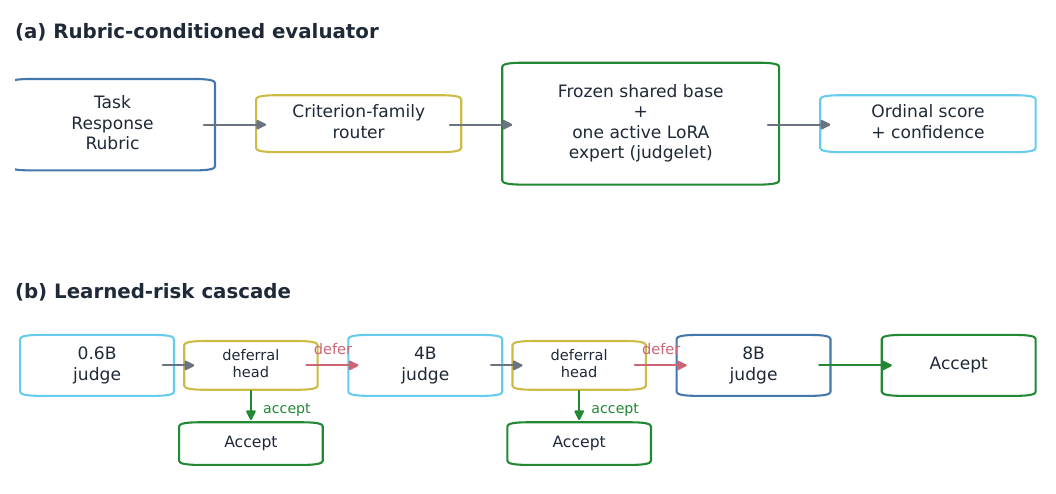}
\caption{Two routing decisions in the evaluation layer. The rubric selects
one criterion-family adapter on a shared base (top). A learned deferral head
then decides whether a reward-model ranking is accepted or forwarded to a
larger evaluator (bottom). Family granularity sets how much supervision is
shared; deferral thresholds are calibrated and audited on disjoint data.}
\label{fig:architecture}
\end{figure}

\section{Related Work}

\paragraph{LLM judges and specialized evaluators.}
Prometheus trains open evaluators conditioned on references and user-defined
score rubrics \citep{kim2023prometheus}; Prometheus 2 extends the approach to
direct assessment and pairwise ranking \citep{kim2024prometheus2}. Small-judge
comparisons find strong domain dependence \citep{laddha2026slmjury}. We start
from that heterogeneity but share one base model across criterion-family
adapters. Quantitative LLM Judges align a judge's score and rationale to
human scores with post-hoc models \citep{sahoo2025quantitative}; our post-hoc
heads predict correctness for routing instead.

\paragraph{Sparse experts and parameter-efficient adaptation.}
LoRA freezes a shared model and expresses an update in low-rank matrices
\citep{hu2022lora}. Sparse mixture-of-experts models grow stored capacity
while activating few parameters per input \citep{fedus2022switch}. Judgelets
borrow both properties at coarser granularity: a rubric-derived router picks
one whole adapter, so the assignment is inspectable and constant for every
example under the same criterion. Mixture-of-LoRA methods report gains from
learned or task-labeled routing \citep{li2024mixlora,feng2024mixture}. Our
controlled result identifies the opposite regime: independently trained
experts that divide one judgment task into small rubric families lose to a
single adapter.

\paragraph{Learning to defer, selective judging, and cascades.}
Deferral has a standard formulation: a system may predict or hand the case to
a downstream decision-maker, trading coverage against risk
\citep{chow1970optimum,geifman2017selective,madras2018predict,
mozannar2020consistent}. Trust or Escalate applies cascaded selective
evaluation with calibrated agreement guarantees to LLM judging
\citep{jung2025trust}; SCOPE studies finite-sample risk control for pairwise
judges \citep{badshah2026scope}; Ask a Strong LLM Judge routes uncertain
reward-model comparisons to a generative judge \citep{xu2025ask}.
Hierarchical abstention and confidence tuning already show that cascades can
improve cost--accuracy tradeoffs
\citep{zellinger2024controlled,rabanser2025gatekeeper,melo2026epistemic}. Our
deferral heads are deliberately plain logistic probes. The question is
narrower and orthogonal: for a fixed evaluation workload, does domain
specialization belong in the judge's parameters or in the deferral rule?

\paragraph{Rubrics and reward-model evaluation.}
AdaRubric generates task-specific rubric dimensions for agent trajectories
\citep{ding2026adarubric}; we instead assume a repeated, versioned collection
of domain rubrics and compile it into fixed system artifacts. RewardBench 2
evaluates factuality, focus, instruction following, mathematics, safety, and
ties \citep{malik2025rewardbench2}. Skywork-Reward-V2 provides a 0.6B-to-8B
ladder trained with one data recipe \citep{liu2026skyworkrewardv2}, so model
size varies while the family stays fixed.\footnote{The exact Qwen3
checkpoints (0.6B, 1.7B, 4B, 8B) are indexed in the official
\href{https://huggingface.co/collections/Skywork/skywork-reward-v2}{Skywork-Reward-V2
collection}.} Cross-family tests use the Tulu-3-8B reward model
\citep{lambert2024tulu3} and an OpenAssistant DeBERTa pair
\citep{openassistantdeberta}. HealthBench supplies physician-written rubric
criteria with physician labels of whether a response meets them
\citep{arora2025healthbench}.

\section{Problem Setting}

Let $\mathcal{R}=\{r_1,\ldots,r_m\}$ be a rubric collection. Each example
contains a task $x$, a candidate response $y$, an optional reference $a$, a
rubric $r$, and an ordinal judgment $z\in\{1,\ldots,L\}$. A monolithic
evaluator predicts $p(z\mid x,y,a,r)$ with one set of trainable parameters.
We instead assign each rubric to a family $g(r)\in\{1,\ldots,K\}$. Family $k$
has a low-rank adapter $\phi_k$ on a frozen shared base $\theta$:
\begin{equation}
p_k(z\mid x,y,a,r)=M_{\theta,\phi_k}(x,y,a,r),\qquad k=g(r).
\end{equation}
Only $\phi_{g(r)}$ is active. The bank stores $K$ adapters and runs one.

Acceptance follows the selective-prediction template. For a confidence
function $c_k$ and threshold $t_k$, family $k$ accepts an automatic judgment
when $c_k\ge t_k$; otherwise the case is deferred. Selective risk and
coverage are
\begin{equation}
R_k(t_k)=\Pr(\hat z\ne z\mid c_k\ge t_k),\qquad
\Gamma_k(t_k)=\Pr(c_k\ge t_k).
\end{equation}
Thresholds are chosen on calibration data and never retuned on development,
test, held-criterion, or external data.

For a cascade of models $M_1,\ldots,M_J$, every example reaches $M_1$ and
deferred examples continue. If $q_j$ is the fraction reaching stage $j$ and
$P_j$ its parameter count, normalized parameter compute is
\begin{equation}
C_{\mathrm{norm}}=\frac{\sum_{j=1}^{J}q_jP_j}{P_J}.
\label{eq:compute}
\end{equation}
This quantity compares architectures. It is a parameter-count proxy, not a
measurement of energy, latency, or cost.

A cascade can beat its terminal judge only through complementary errors. For
two stages, let $A$ denote early acceptance and $C_j$ correctness of stage
$j$. Relative to always using stage 2,
\begin{equation}
\Delta_{\mathrm{acc}}=
\Pr(A\cap C_1\cap\neg C_2)-
\Pr(A\cap\neg C_1\cap C_2).
\label{eq:rescue}
\end{equation}
We call the first term \emph{rescue} and the second \emph{harm}: the rates at
which early acceptance recovers small-model disagreements that help and
keeps ones that hurt. Routing pays only if rescue exceeds harm and the first
stage is cheap enough that $P_1+(1-\Pr(A))P_2<P_2$. The decomposition
separates ordinary selective prediction from the stronger claim, tested
below, that a cascade improves accuracy and compute at once.

\section{Rubric Compilation and Deferral}

\subsection{Frozen criterion splits}

A rule is identified by its rubric text, so we hold out whole criterion
strings. A deterministic hash reserves a set of them for
\texttt{shift\_locked}, an unseen-criterion split kept out of every
training, calibration, development, and test decision. For the remaining criteria, all responses that share a
criterion--instruction pair land in the same split, so related responses to
one task never cross partitions. The test split is frozen before any
analysis; we refer to it as the locked test.

\subsection{Criterion-family router and judgelet bank}

The compiler represents criterion strings with word and bigram TF--IDF
features fitted only on criteria available to adapter training. K-means then
defines $K$ criterion families; new rubrics are assigned by the fitted
vectorizer and centroids. The router is deliberately simple. Its assignments
can be read directly, and routing quality is not the variable under study.

We evaluate $K\in\{1,4,8\}$, where $K=1$ is the monolithic evaluator. All
judgelets share the 0.6B Skywork-Reward-V2 Qwen3 base. We replace its scalar
reward head with a five-class score head and train rank-8 LoRA modules on
attention and feed-forward projections. Each judgelet has 5.05M trainable
parameters, 0.84\% of the full model. A rank-64 monolithic
adapter matches the aggregate stored capacity of eight rank-8 judgelets.

The primary bank matches the monolith's total optimizer budget: its eight
experts receive 98--489 updates (1,494 in all), against 1,489 monolithic
updates. Matched totals do not mean every expert converged, so an
optimization control cycles every family to exactly 500 updates (4,000
total). A second bank starts each expert from the trained rank-8 monolith
and runs one family epoch at learning rate $5\times10^{-5}$, preserving
shared training before the split. Because the primary scratch bank uses
$2\times10^{-4}$, we also train a scratch bank at $5\times10^{-5}$ for the
same 1,494 family updates. That control isolates initialization within the
family-adaptation recipe; it is not a learning-rate sweep.

Inputs place the criterion and its five score descriptions before the task,
response, and reference. A full-corpus tokenizer audit found that over 95\%
of inputs exceed 512 tokens and under 3.5\% exceed 1,024. Primary models use
1,024 tokens. A 512-token variant serves as a truncation intervention, and
the rubric-ablated task-and-response control uses 512 tokens.

\subsection{Risk-controlled acceptance}

We compare one global confidence threshold, one threshold per criterion
family, and a family policy with a Bonferroni correction across the $K$
families. Each threshold comes from scanning the calibration risk--coverage
curve. Because the scan is adaptive, the thresholds are calibration
heuristics, not certificates: realized risk is evaluated on untouched splits
with one-sided 95\% Wilson upper bounds.

\subsection{Learned deferral heads}

Reward-model margins are not calibrated comparably across skills. For each
non-terminal model we fit a logistic \emph{deferral head} that predicts
whether the stage's top-ranked candidate is correct---a correctness probe in
the learning-to-defer sense \citep{madras2018predict,mozannar2020consistent}.
Its inputs are the stage's own score geometry (largest scores, gap, range,
mean, variance, candidate count, winning position) and a one-hot skill
family. Fitting, threshold selection, and evaluation use three disjoint
partitions. The head never edits reward scores; it accepts the current
ranking or forwards the example.\footnote{With the five non-tie RewardBench
2 skills, the head has 14 inputs and one intercept. Its arithmetic is
negligible next to a 0.6B forward pass and is excluded from
Eq.~\ref{eq:compute}; every reward-model forward pass is included.}

We require a 3\% one-sided upper bound on calibration error, leaving a
declared buffer under the 5\% held-out target. The untouched test partition
is checked with a Wilson diagnostic and an exact one-sided 95\%
Clopper--Pearson bound. As a stage-selection rule fixed on calibration data,
any non-terminal stage that accepts no calibration example is dropped. Raw
margin, logistic margin calibration, isotonic margin calibration, and a
geometry-only probe (no skill indicator) use the same partitions; the last
tests whether domain identity adds signal beyond score geometry.

\section{Experimental Protocol}

The direct-assessment corpus contains 99,952 Feedback Collection examples
and 996 criteria.\footnote{Public releases:
\href{https://huggingface.co/datasets/prometheus-eval/Feedback-Collection}{Feedback
Collection} and
\href{https://huggingface.co/datasets/prometheus-eval/Feedback-Bench}{Feedback-Bench}.
The artifact manifest records row counts and SHA-256 hashes for the
immutable files used here.} Criterion holdout reserves 206 criteria and
20,696 examples. The remainder gives 47,638 training, 9,441 calibration,
9,597 development, and 12,580 locked-test examples. Feedback-Bench adds
1,000 external examples; all 626 of its criterion strings occur in Feedback
Collection, but 122 were excluded from adapter training and account for 203
external examples.

RewardBench 2 contains 1,865 tasks and 8,977 candidate
completions.\footnote{Official test release:
\href{https://huggingface.co/datasets/allenai/reward-bench-2}{allenai/reward-bench-2};
its 102 tie tasks stay in the official macro metric.} We reproduce the
official subset metric on all tasks. The cascade study uses the 1,763
non-tie tasks; tie tasks are excluded from head fitting, threshold
calibration, and binary cascade accuracy.

The real-rubric study uses the HealthBench meta-evaluation release
\citep{arora2025healthbench}: 29,511 triples of a health conversation, a
model response, and a physician-written rubric criterion, each labeled by
two or more physicians as met or not met.\footnote{File
\texttt{2025-05-07-06-14-12\_oss\_meta\_eval.jsonl} from the official
\texttt{simple-evals} release.} The criteria and labels are real physician
work; the conversations themselves are benchmark material, produced by
synthetic generation and adversarial human writing rather than drawn from
patient records. We keep the 23,486 rows with a strict
physician majority (99.1\% of them unanimous) and drop ties. The label is
binary, so the judge gets a two-class head; base model, LoRA rank, learning
rate, epoch count, and token limit match the Prometheus monolith. The 30
consensus criteria are the rubric collection. A deterministic hash holds
out 9 complete criteria (8,466 rows) as the unseen-criterion split; the
rest is split by conversation into 9,019 training, 1,807 calibration, 1,760
development, and 2,434 locked-test rows. Because criteria are held out
before conversations are hashed, a conversation in the unseen-criterion
split can also appear, under different criteria, in training (655 of its
1,516 conversations do); Section~\ref{sec:healthbench} therefore reports
the conversation-disjoint subset as well. The counterfactual copy swaps the
criterion of every test and unseen-criterion row for one from a different
criterion category. Paired comparisons resample by conversation.

Direct-assessment metrics are accuracy, macro F1, mean absolute score error
(MAE), quadratic-weighted kappa, Spearman correlation, calibration error,
selective coverage and risk, and one-sided 95\% Wilson bounds. Comparisons
use paired bootstrap intervals with criterion ID as the resampling unit. We
report active and stored adapter parameters, wall-clock training time,
checkpoint size, and inference throughput.

The rubric intervention replaces the criterion and all five score
descriptions with a rubric drawn from a different cluster in the same split.
Task, response, reference, label, router assignment, and row identity stay
fixed. A model that ignores the rubric is invariant to the swap; a
rubric-sensitive evaluator should not be.

A content-hash audit found two duplicated pairs (four rows of 99,952) with
conflicting labels. Neither pair crosses a split, criterion, or routed
family. All joins and paired tests use the unique row identifier.

\section{Results}

\subsection{The rubric helps, but the response is a strong shortcut}

On the locked test, the full-context monolith beats the rubric-ablated
control by 2.11 accuracy points and lowers MAE by 0.030
(Table~\ref{tab:direct}). A paired bootstrap over 765 criterion IDs gives
$+1.48$ to $+2.75$ points.\footnote{Throughout, ``points'' are
percentage-point differences and intervals are 95\% intervals.} On held
criteria the gap grows to $+3.10$ points ($+2.47$ to $+3.74$).

\begin{table}[t]
\caption{Direct-assessment results. Higher is better except MAE. ``Shift''
is the unseen-criterion split; ``External'' is Feedback-Bench.}
\label{tab:direct}
\centering
\small
\begin{tabular}{lrrrrrr}
\toprule
Model & Dev & Test & MAE & $\kappa$ & Shift & External\\
\midrule
Rubric-ablated, 512 & 74.75 & 74.27 & .281 & .916 & 74.75 & 75.70\\
Monolithic rubric, rank 8 & 76.78 & 76.38 & .250 &
.929 & 77.85 & 77.60\\
Monolithic rubric, rank 64 & \textbf{79.09} & \textbf{77.50} &
\textbf{.237} & \textbf{.934} & \textbf{78.97} & \textbf{82.10}\\
\bottomrule
\end{tabular}
\end{table}

The external comparison is less decisive: the 1.9-point accuracy gain has a
criterion-bootstrap interval of $-0.5$ to $+4.4$ ($p=0.138$), though MAE
falls from 0.270 to 0.239 ($p=0.033$). Full-context accuracy is 79.30\% on
the 797 external examples whose criteria were available to adapter training
and 70.94\% on the 203 governed by held-out criteria. Raising monolithic
rank from 8 to 64 adds 1.11 test points ($+0.54$ to $+1.71$), 1.12 shift
points, and 4.50 external points.

Context length matters at inference time. Running the same adapter with a
512-token limit drops test accuracy to 73.82\%, a $-2.56$-point paired
difference ($-3.16$ to $-1.97$), and raises MAE by 0.038. That is
indistinguishable from the rubric-ablated control ($-0.45$ points; $-1.06$
to $+0.15$; $p=0.147$), even though throughput roughly doubles.
Risk-controlled coverage under the Bonferroni policy falls from 24.44\% to
19.77\%. Truncate the rubric-bearing context and the rubric benefit is gone.

\subsection{Counterfactual rubrics remove the gain}

Replacing the rubric while holding everything else fixed lowers locked-test
accuracy from 76.38\% to 73.72\%, a paired difference of $-2.66$ points
($-3.26$ to $-2.09$). On the unseen-criterion split, accuracy falls from
77.85\% to 75.34\% ($-3.05$ to $-1.97$). The wrong-rubric condition also
trails the rubric-ablated control on the locked test, which rules out an
account based on extra benign context.

\subsection{Global calibration overstates safe coverage}

The global threshold accepts 37.47\% of locked-test examples, but its 5.38\%
risk upper bound misses the 5\% target (Table~\ref{tab:selective}). Family
thresholds pass at 32.06\% coverage (4.58\% bound). The Bonferroni policy is
stricter: 24.44\% coverage, 3.54\% bound. On the unseen-criterion split it
covers 26.13\% with a 3.05\% bound.

\begin{table}[t]
\caption{Locked-test acceptance with calibration-only thresholds.}
\label{tab:selective}
\centering
\small
\begin{tabular}{lrrrr}
\toprule
Evaluator and policy & Coverage & Risk & Upper & Pass\\
\midrule
Rubric-ablated, global & 29.93 & 4.91 & 5.53 & No\\
Rubric-ablated, family Bonf. & 18.01 & 2.69 & 3.31 & Yes\\
Rubric, global & 37.47 & 4.84 & 5.38 & No\\
Rubric, family & \textbf{32.06} & 4.04 & 4.58 & Yes\\
Rubric, family Bonf. & 24.44 & 2.99 & 3.54 & Yes\\
\bottomrule
\end{tabular}
\end{table}

With a wrong rubric, the family policy's test risk bound climbs to 5.30\%
while coverage drops to 27.24\%. The Bonferroni policy stays under target
but accepts only 20.95\%. An acceptance rule audited with valid rubric text
cannot be assumed safe once that text is corrupted.

\subsection{Physician-labeled rubrics reproduce the mechanism}
\label{sec:healthbench}

Feedback Collection is synthetic end to end, labels included. The rubric
effect could simply reflect how that supervision was generated. The
HealthBench study swaps in real expert work: physicians wrote the criteria
and the labels, applied to realistic benchmark conversations
(Table~\ref{tab:healthbench}).

On the locked test, the rubric-conditioned judge reaches 88.41\% accuracy
and 0.727 macro F1; the rubric-ablated control reaches 86.81\% and 0.648.
The paired difference is $+1.60$ points (conversation-clustered interval
$+0.46$ to $+2.74$, $p=0.004$), and the gap in balanced accuracy is wider,
68.7\% against 61.5\%, because the ablated judge leans harder on the
84.2\%-positive base rate. Swapping in a criterion from a different
category costs $4.31$ points ($-5.57$ to $-3.05$, $p=0.0002$) and lands the
judge at the always-met baseline. Physician-written rule text is doing the
work; the response alone does not explain the score.

\begin{table}[t]
\caption{HealthBench physician meta-evaluation. ``Wrong rubric'' swaps the
criterion at inference time; ``always met'' predicts the majority class.
Balanced accuracy is the mean of the two class recalls.}
\label{tab:healthbench}
\centering
\small
\begin{tabular}{lrrrrrr}
\toprule
& \multicolumn{3}{c}{Test (21 seen criteria)} &
\multicolumn{3}{c}{Unseen criteria (9)}\\
\cmidrule(lr){2-4}\cmidrule(lr){5-7}
Condition & Acc. & Bal. & F1 & Acc. & Bal. & F1\\
\midrule
Rubric-conditioned & \textbf{88.41} & \textbf{68.7} & \textbf{.727}
& 82.92 & 55.1 & \textbf{.556}\\
Rubric-ablated & 86.81 & 61.5 & .648 & 82.84 & 54.1 & .540\\
Wrong rubric & 84.10 & 61.4 & .634 & 80.64 & 54.3 & .546\\
Always met & 84.22 & 50.0 & .457 & \textbf{83.33} & 50.0 & .455\\
\bottomrule
\end{tabular}
\end{table}

The unseen-criterion split gives the sharper lesson. Against the ablated
control the rubric adds nothing there ($+0.08$ points; $-0.34$ to $+0.51$;
$p=0.70$), and neither condition beats the always-met baseline on raw
accuracy. Yet corrupting the rubric still costs $2.28$ points ($-2.80$ to
$-1.77$, $p=0.0002$): the judge reads the text even when reading fails to
pay. Restricting to the 5,501 rows whose conversations never appear in
training changes nothing: $+0.06$ points ($-0.51$ to $+0.60$, $p=0.86$)
against the ablated control, and $-2.36$ ($-3.02$ to $-1.72$) under a wrong
rubric. Conversation overlap does not explain the null. The contrast
with Prometheus, where 790 training criteria supported a $+3.10$-point gain
on held-out rules, points to rule breadth: 21 distinct training criteria
are too few to learn rule-reading that transfers. Breadth is a different
axis from the per-family example depth of Section~\ref{sec:scaling}, and
this study does not vary the two independently.

Acceptance thresholds inherit the same boundary. A calibration-selected
global threshold accepts 22.19\% of test rows at 3.89\% realized risk, just
missing the 5\% Wilson bound (5.50\%); the ablated judge accepts nothing.
Applied to unseen criteria, the same threshold keeps 22.6\% coverage but
realized risk rises to 8.73\%. A threshold audited on familiar rules is not
safe on unfamiliar ones, which is the deployment-facing version of the
criterion-shift warning from the Prometheus study.

\subsection{Deferral heads change the accuracy--compute frontier}

The four Skywork models score 61.39, 68.11, 75.44, and 77.91 on the official
RewardBench 2 macro metric as size grows from 0.6B to 8B. The cascade
analysis uses a different number: micro top-choice accuracy on the 1,763
non-tie tasks. The two are not directly comparable. Over 20
train/calibration/test repartitions, the best cascade forwards an average of
19.80\% of test tasks to the 8B terminal stage and reaches 89.40\% accuracy
at 0.415 normalized compute, 4.66 points above 8B alone on the same
partitions (Table~\ref{tab:cascade}).

The gain ranges from 3.21 to 6.19 points across the 20 deterministic
repartitions. Because the underlying tasks are reused, that is a stability
range, not an independent-sample confidence interval. Complementary errors
make the gain possible: an oracle over all four sizes reaches 91.80\%, and
one restricted to the deployed 0.6B--4B--8B subset reaches 90.97\%. Against
8B alone, early acceptance rescues 5.32\% of tasks and harms 0.66\%, the
$+4.66$-point net in Eq.~\ref{eq:rescue}.

\begin{table}[t]
\caption{RewardBench 2 routing on 1,763 non-tie tasks, mean over 20 frozen
repartitions (mean test size 616.8). ``Exact pass'' counts runs whose
one-sided 95\% Clopper--Pearson bound on early-acceptance error is at most
5\%. Compute is normalized to running 8B on every task.}
\label{tab:cascade}
\centering
\small
\begin{tabular}{lrrrr}
\toprule
Policy and stages & Early cov. & Accuracy & Exact pass & Compute\\
\midrule
8B only & 0.00 & 84.75 & -- & 1.000\\
Raw margin, 4B--8B & 55.61 & 84.81 & 20/20 & 0.944\\
Isotonic margin, 4B--8B & 48.86 & 84.84 & 19/20 & 1.011\\
Geometry probe, 0.6B--4B--8B & 80.35 & 89.37 & 20/20 & 0.413\\
Deferral head, 0.6B--8B & 71.52 & 88.71 & 20/20 & \textbf{0.360}\\
Deferral head, 0.6B--4B--8B & \textbf{80.20} & \textbf{89.40} &
\textbf{20/20} & 0.415\\
Deferral head, all four stages & 80.20 & 89.40 & 20/20 & 0.476\\
\bottomrule
\end{tabular}
\end{table}

Logistic calibration preserves the raw-margin ordering, so threshold search
lands on the same operating point. Isotonic calibration adds ties and no
accuracy. The learned head behaves differently: it uses score geometry and
skill identity to find cases where a smaller model is right, including some
the 8B model gets wrong, so the terminal is not treated as an oracle.
Dropping the skill one-hot changes accuracy by 0.03 points and compute by
$-0.002$; most of the signal is multivariate, stage-local score geometry
rather than the skill label.

\begin{figure}[t]
\centering
\includegraphics[width=\linewidth]{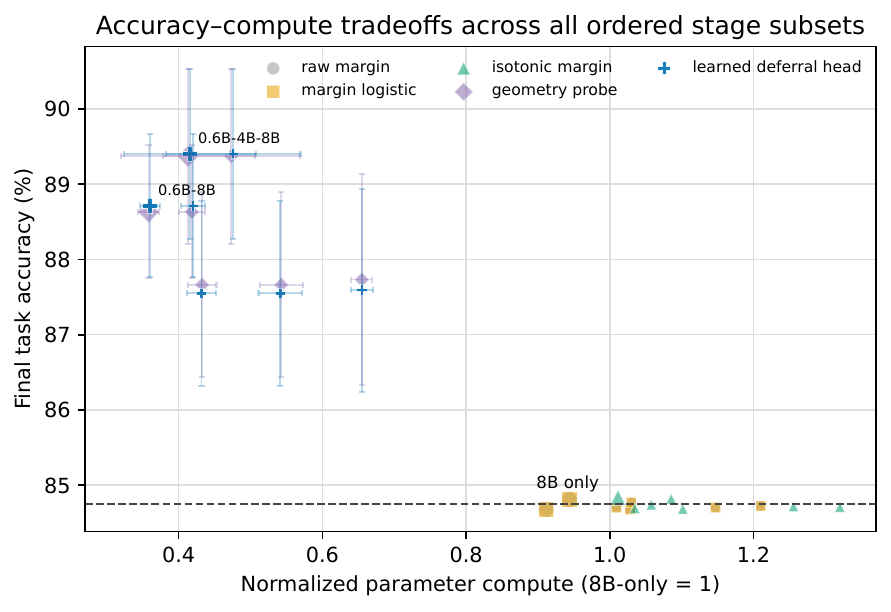}
\caption{All seven compute-ordered stage subsets and five confidence
policies. Learned deferral forms its own accuracy--compute frontier.
Removing the 1.7B stage keeps 89.40\% accuracy and cuts normalized compute
from 0.476 to 0.415. Marker shapes identify policies; error bars on learned
probes show one standard deviation across repartitions.}
\label{fig:pareto}
\end{figure}

The subset comparison explains why the first four-stage result misled. The
1.7B model accepted no calibration example in any run, so the
calibration-fixed pruning rule removes it before locked evaluation. It would
have accepted no test example either, while every example paid its forward
pass. The 0.6B--8B pair is the minimum-compute point at 0.360; adding 4B
costs 0.055 and buys 0.69 accuracy points. Stage membership is part of the
fitted system, not an implementation detail.

Not every judge pair should cascade. Composing the Prometheus rank-8 and
rank-64 adapters accepts 27.98\% of test examples early and moves accuracy
from 77.50\% to 77.52\%, at 1.720 forward passes normalized to rank-64
alone. The two adapters share a 0.6B base and nearly the same inference
cost. Complementary errors are not enough when the first stage costs
nearly as much as the second.

\subsection{Cascades across model families}
\label{sec:crossfamily}

So far every cascade stays inside one model family. The gain might then
rest on complementary errors specific to one training recipe. Two transfers
test that directly (Table~\ref{tab:crossfamily}).

The first swaps the terminal. The Tulu-3-8B reward model
\citep{lambert2024tulu3} shares neither base (Llama~3.1 versus Qwen3) nor
data recipe with the Skywork ladder, and it is weaker: 59.02 on the
official macro metric against 77.91, or 64.58\% terminal-only accuracy on
the cascade protocol. The deferral heads do not change: each is fitted to
its own stage's correctness and never consults the terminal. The acceptance
decisions are therefore identical to the Skywork-terminal runs; only the
fate of forwarded examples changes. In front of the weaker terminal, the
same 0.6B--4B early stages rescue 22.0\% of tasks. The system lands at
86.43\%, 21.85 points above the terminal alone, at the same 0.415 compute,
and the exact audit passes 20/20. A weaker terminal means more rescue, not
a broken protocol. Raw margins, by contrast, transfer badly: the
0.6B-margin cascade reaches 65.50\% at 1.009 compute and passes 3/20.

The second transfer changes the first stage's family in front of the strong
terminal. A 184M OpenAssistant DeBERTa-base first stage, whose own official
RewardBench 2 score is only 26.07, accepts 30.1\% of tasks ahead of
Skywork-8B and lifts it from 84.75\% to 87.95\% at 0.723 compute (exact
audit 19/20; the geometry-only probe reaches 88.00\% and passes 20/20). Margin rules accept
nothing in this pairing: DeBERTa margins fail the 3\% calibration bound
outright, which is what motivated learned heads in the first place.

\begin{table}[t]
\caption{Cross-family cascades with learned deferral heads, mean over the
same 20 frozen repartitions. Terminal and cascade accuracies are micro
top-choice accuracy on non-tie tasks; compute is normalized to running the
listed terminal on every task. The first two rows repeat the same-family
reference.}
\label{tab:crossfamily}
\centering
\small
\begin{tabular}{llrrrr}
\toprule
First stage(s) & Terminal & Term. acc. & Cascade & Exact pass & Compute\\
\midrule
Skywork 0.6B & Skywork 8B & 84.75 & 88.71 & 20/20 & 0.360\\
Skywork 0.6B, 4B & Skywork 8B & 84.75 & \textbf{89.40} & 20/20 & 0.415\\
\midrule
Skywork 0.6B & Tulu-3 8B & 64.58 & 82.35 & 20/20 & \textbf{0.360}\\
Skywork 0.6B, 4B & Tulu-3 8B & 64.58 & 86.43 & 20/20 & 0.415\\
Skywork 4B & Tulu-3 8B & 64.58 & 88.62 & 19/20 & 0.655\\
OA DeBERTa base & Skywork 8B & 84.75 & 87.95 & 19/20 & 0.723\\
OA DeBERTa base & OA DeBERTa large & 40.14 & 48.78 & 19/20 & 1.122\\
\bottomrule
\end{tabular}
\end{table}

The last row shows the constraint that does bind. With OA-large as its own
terminal, the same DeBERTa-base first stage reproduces the accuracy effect
(40.14\% to 48.78\%) but costs 1.122 compute, because the 184M base is
already 42\% the size of the 435M terminal. In the two pairings tested,
family membership was not what mattered; relative price was. The design
rule needs the early judge to be cheap and its errors complementary, and
both cross-family directions satisfied it when the size ratio did. Both
results remain within RewardBench 2 and a parameter-count cost proxy.

\subsection{Fine-grained specialization fragments supervision}

The eight-family bank is much worse than the monolithic evaluator:
locked-test accuracy 66.34\%, 10.05 points below the rank-8 monolith
($-10.87$ to $-9.22$), with MAE up from 0.250 to 0.375. The held-criterion
gap is $-10.13$ points ($-11.14$ to $-9.13$). Expert quality is uneven: a
few reach 71--72\%, several sit between 56\% and 61\%.

Neither global nor family calibration passes the bank's locked-test risk
audit. Bonferroni calibration passes at a 4.87\% bound with 5.43\% coverage,
against 24.44\% for the monolith. Counterfactual rubric replacement costs a
further 1.72 points ($-2.39$ to $-1.07$), so the bank is not ignoring the
rubric. It has simply learned the common scoring task less well once the
supervision was divided.

Four families narrow the loss without removing it: 71.81\% on the locked
test, 4.57 points below the monolith ($-5.26$ to $-3.89$), 73.78\% on held
criteria. Under independent one-epoch training the pattern in
Table~\ref{tab:granularity} is monotone: active size fixed, stored
parameters up, accuracy down as the data are split more finely. The
four-family Bonferroni policy passes the locked test at 17.79\% coverage,
6.65 points below monolithic coverage, and just misses on the
unseen-criterion split with a 5.006\% bound.

Extra optimization recovers only part of the gap. Cycling each expert to 500
updates lifts the bank from 66.34\% to 72.56\%. The 6.22-point gain costs
4,000 updates and still sits 3.82 points under the rank-8 monolith and 4.94
under rank 64. Under-training matters; it is not the explanation.

Shared pre-adaptation is far more effective. Starting each expert from the
trained rank-8 monolith and running one family epoch reaches 76.85\% test
accuracy: 10.51 points above independent training and 0.47 above the
monolithic initialization itself ($+0.13$ to $+0.82$). Held-criterion
accuracy is 78.14\%; external accuracy is 78.90\%. A wrong rubric still
costs 3.28 points ($-3.90$ to $-2.66$), so the recovery does not come from
ignoring the family-specific rule text. The warm-started bank does receive
more total optimization than the monolith it starts from; the budget sweep
in Section~\ref{sec:scaling} adds the control that separates that from
specialization.

The learning-rate-matched scratch bank sharpens the comparison. At the same
$5\times10^{-5}$ rate, one family epoch, and 1,494 aggregate updates,
scratch training reaches 56.91\% on the locked test, 58.47\% on held
criteria, 58.00\% externally. Shared initialization is 19.94 points better
on the locked test ($+18.88$ to $+21.02$). The low-rate scratch run clearly
underfits, so it says nothing about whether $5\times10^{-5}$ is a good
scratch rate in general. It does rule out the claim that shared
initialization wins merely by using that rate. Together with the
4,000-update control, the experiments separate shared pre-adaptation from
family step count and family learning rate.

On the external set, the fixed-update control ends 0.50 points above shared
initialization (79.40\% versus 78.90\%), with a criterion-bootstrap interval
of $-2.65$ to $+1.70$. That comparison is unresolved; it is not a reversal
of the locked-test result.

The selective audit draws the same line. The shared-initialization bank
accepts 24.90\% of locked-test examples under a 3.48\% one-sided bound. The
fixed-update bank selects 11.76\% but misses with a 5.36\% bound; the
low-rate scratch bank accepts nothing.

Capacity points the other way entirely. Rank 64 stores 40.38M adapter
parameters, almost exactly the bank's 40.41M, yet scores 11.16 test points
higher ($+10.28$ to $+12.07$), exceeds rank 8 by 1.11 points, and reaches
30.14\% Bonferroni coverage. Stored capacity is not the limit. The optimization
controls instead point to the loss of common judgment training when
supervision is fragmented, and shared pre-adaptation nearly closes the gap.

\subsection{How much data justifies a split?}
\label{sec:scaling}

\begin{figure}[t]
\centering
\includegraphics[width=0.7\linewidth]{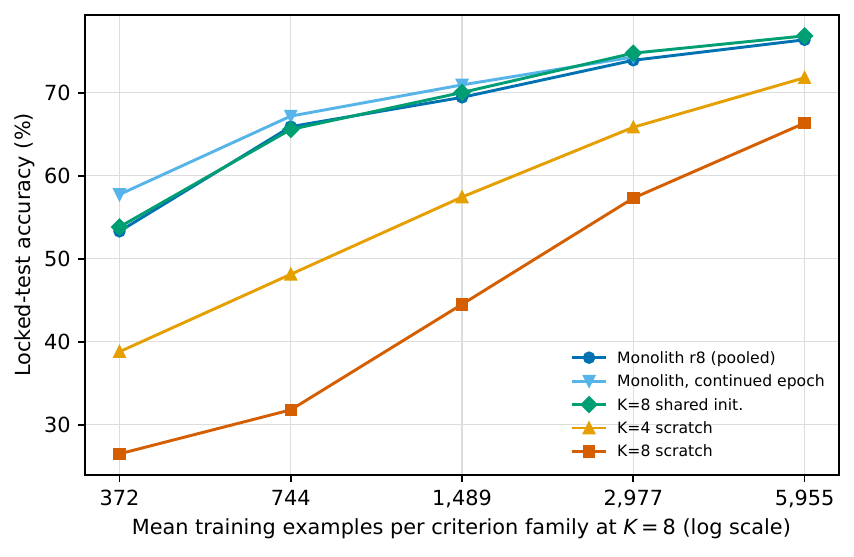}
\caption{Locked-test accuracy as the training budget grows, by design.
Scratch fragmentation never reaches the monolith at either granularity, and
its cost grows with $K$. The warm-started split tracks the monolith but
stays under the continued-pooled control until the control itself stops
improving, near 3,000 examples per family.}
\label{fig:budget}
\end{figure}

The original comparison fixes the corpus and varies the design. ``Enough
data to justify a split'' deserves numbers. We therefore retrain four
designs at four nested budgets, from 1/16 to 1/2 of the corpus (372 to
2,977 examples per family at $K{=}8$, on average), with recipes, seeds, and
evaluation splits unchanged. The designs are the rank-8 monolith, scratch
banks at $K{=}4$ and $K{=}8$, and the $K{=}8$ shared-initialization bank. A
fifth run is the matched-optimization control. It takes each budget's monolith and continues
pooled training for one more epoch at $5\times10^{-5}$: the same extra
optimization the warm-started bank receives, spent on pooled rather than
family-split data. Full-data points complete the curves
(Figure~\ref{fig:budget}; Appendix Table~\ref{tab:scaling}).

Scratch specialization loses at every scale and both granularities. At
$K{=}8$ the deficit is 26.8 points at 372 examples per family. It peaks at
34.1 at 744, where pooled data lifts the monolith by 12.6 points but eight
independent experts gain only 5.3. It then narrows steadily: 24.9 at 1,489,
16.6 at 2,977, 10.05 at 5,955. Halving the granularity roughly halves the
damage without changing its shape; the $K{=}4$ banks trail by 14.5, 17.8,
12.0, 8.1, and 4.6 points across the same budgets. The last three $K{=}8$
gaps close at about 6--8 points per data doubling. At that rate, break-even
sits past 12,000--24,000 examples per family, two to four times the corpus.
That is an extrapolation, not an observation. Fragmentation cost grows with
$K$ and shrinks with data, and nothing in this range comes close to paying.

Shared initialization removes most of that cost, but below saturation not
all of it. Next to the same-budget monolith, the warm-started bank looks
free: within 0.9 points everywhere, and slightly ahead from 1,489 examples per
family on. The matched control shows what that comparison hides. One more
pooled epoch beats the same epoch spent family-split: by 3.91 points at 372
examples per family ($[-4.57,-3.24]$), by 1.61 at 744 ($[-2.14,-1.06]$),
and by 0.91 at 1,489 ($[-1.40,-0.45]$). Only at 2,977 does the order flip.
There the split edges out continued pooling by 0.46 points
($[+0.03,+0.89]$), just where continued pooling stops paying ($+0.40$ over
the monolith, $[-0.02,+0.82]$). The rule this gives is sharper than an
example count. Keep pooling while another pooled epoch still pays. Split,
warm-started, once it does not; here that point sits near 3,000 examples
per family. Each grid cell is a single training run, so the exact crossover
deserves more seeds before it is read as a constant.

Risk-controlled coverage has its own, later threshold. Below 1,489 examples
per family no design accepts anything at the 5\% bound. At 1,489 the
monolith accepts 1.9\% with a passing 4.8\% upper bound, while the shared
bank's 1.5\% acceptance misses the bound (5.2\%). At 2,977 both accept
about 15\% and pass; at full data, 24.4\% and 24.9\%. The scratch banks
accept nothing at any reduced budget. Autonomy, in other words, trails
accuracy: acceptance at useful coverage arrives around the same $\sim$3,000
examples per family as the split crossover, and accepting a quarter of the
workload needs the full corpus.

\begin{figure}[t]
\centering
\includegraphics[width=0.82\linewidth]{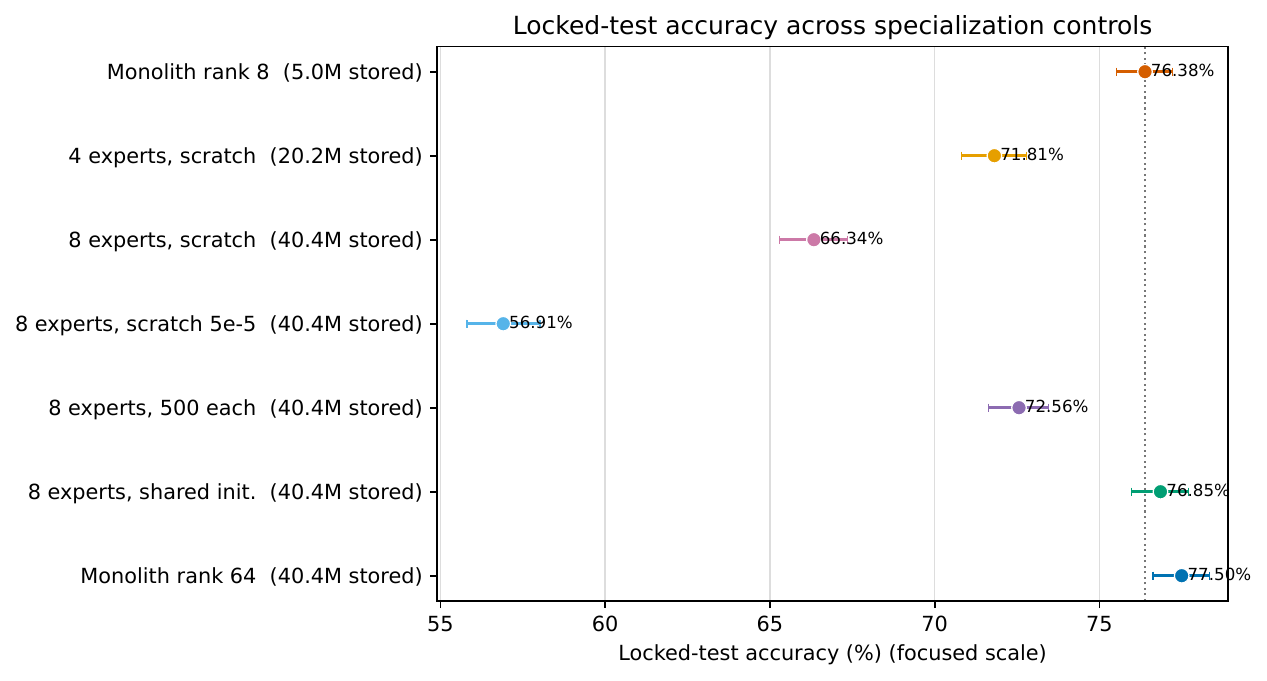}
\caption{Locked-test accuracy across specialization and optimization
controls. Dots show criterion-bootstrap 95\% intervals; the focused
horizontal scale is used for point estimates rather than bar lengths.}
\label{fig:fragmentation}
\end{figure}

\section{Limitations}

Feedback Collection is synthetic, and a model can exploit its regularities
without reading the rubric; the strong rubric-ablated control shows this is
a property of the data. The HealthBench study answers the objection partway:
physician-written criteria and physician labels reproduce the rubric effect
and the counterfactual sensitivity. Its conversations, however, are
benchmark material, synthetically generated or adversarially written rather
than taken from patient records. The study validates the supervision, not
the deployment distribution. It also remains one domain, with binary rather
than five-point labels, and its 30 consensus criteria are fewer and broader
than Feedback Collection's 996; with only 21 of them in training,
rule-reading did not transfer to the held-out rules there.

Criterion holdout tests unseen rubric text. It does not simulate a policy
revision over time, an adversarial rule, or a shift in the population of
actors. Feedback-Bench is external at the example level, yet all of its
criteria occur somewhere in Feedback Collection; we therefore report the
adapter-held subset and avoid calling the benchmark wholly new-criterion.

The cascade evidence now spans three base-model families (Qwen3, Llama~3.1,
DeBERTa) and three training recipes, but reward-model routing is still
evaluated on repeated partitions of one benchmark, thresholds are audited
retrospectively, and parameter count remains a compute proxy. Appendix~\ref{sec:add-limitations} states the
boundaries in detail.

\section{Conclusion}

The experiments separate two kinds of specialization that usually travel
together. Specializing judge weights too early divides the examples needed
to learn a shared scoring task; more family updates recover part of the
loss, and shared pre-adaptation recovers most of it. The budget grid makes
the rule concrete. Scratch specialization trails the monolith at every
budget and both granularities tested, by 5 to 34 points, and halving $K$
roughly halves the gap; break-even extrapolates, unobserved, past 12,000
examples per family. A warm-started split avoids that cost against a fixed
monolith, but the matched continued-training control shows pooling still
wins below roughly 3,000 examples per family. The split becomes free
exactly where another pooled epoch stops paying.
Risk-controlled autonomy needs more data still before any design accepts a
useful fraction of its own judgments.

Specializing the deferral decision behaves differently. A small correctness
head exploits complementary errors using only the current stage's scores.
The resulting cascade improves accuracy and compute together whenever the
earlier evaluators are genuinely cheap: within the Skywork ladder, across
families with a Tulu terminal, and with a DeBERTa first stage. It fails for
the Prometheus adapter pair, which shares one base and nearly one price.
The design rule is conditional, not universal. Pool judgment learning by
default. Specialize the deferral rule once its risk and cost have been
measured. And let the marginal value of continued pooled training, not a
raw example count, decide when weight specialization stops costing
anything.

\bibliography{references}

@inproceedings{amdahl1967validity,
  title     = {Validity of the Single Processor Approach to Achieving Large Scale Computing Capabilities},
  author    = {Amdahl, Gene M.},
  booktitle = {Proceedings of the April 18--20, 1967, Spring Joint Computer Conference},
  pages     = {483--485},
  year      = {1967},
  doi       = {10.1145/1465482.1465560}
}

@inproceedings{kim2023prometheus,
  title     = {Prometheus: Inducing Fine-Grained Evaluation Capability in Language Models},
  author    = {Kim, Seungone and Shin, Jamin and Cho, Yejin and Jang, Joel and Longpre, Shayne and Lee, Hwaran and Yun, Sangdoo and Shin, Seongjin and Kim, Sungdong and Thorne, James and Seo, Minjoon},
  booktitle = {International Conference on Learning Representations},
  year      = {2024}
}

@inproceedings{kim2024prometheus2,
  title     = {Prometheus 2: An Open Source Language Model Specialized in Evaluating Other Language Models},
  author    = {Kim, Seungone and Suk, Juyoung and Longpre, Shayne and Lin, Bill Yuchen and Shin, Jamin and Welleck, Sean and Neubig, Graham and Lee, Moontae and Lee, Kyungjae and Seo, Minjoon},
  booktitle = {Proceedings of the 2024 Conference on Empirical Methods in Natural Language Processing},
  pages     = {4334--4353},
  publisher = {Association for Computational Linguistics},
  year      = {2024},
  doi       = {10.18653/v1/2024.emnlp-main.248}
}

@article{laddha2026slmjury,
  title   = {{SLMJury}: Can Small Language Models Judge as Well as Large Ones?},
  author  = {Laddha, Anish and Pradhan, Nitesh and Srivastava, Gaurav},
  journal = {arXiv preprint arXiv:2606.07810},
  year    = {2026}
}

@inproceedings{malik2025rewardbench2,
  title     = {{RewardBench 2}: Advancing Reward Model Evaluation},
  author    = {Malik, Saumya and Pyatkin, Valentina and Land, Sander and Morrison, Jacob and Smith, Noah A. and Hajishirzi, Hannaneh and Lambert, Nathan},
  booktitle = {International Conference on Learning Representations},
  year      = {2026}
}

@inproceedings{jung2025trust,
  title     = {Trust or Escalate: {LLM} Judges with Provable Guarantees for Human Agreement},
  author    = {Jung, Jaehun and Brahman, Faeze and Choi, Yejin},
  booktitle = {International Conference on Learning Representations},
  year      = {2025}
}

@inproceedings{xu2025ask,
  title     = {Ask a Strong {LLM} Judge when Your Reward Model is Uncertain},
  author    = {Xu, Zhenghao and Lu, Qin and Zhang, Qingru and Qiu, Liang and Hong, Ilgee and Yu, Changlong and Yao, Wenlin and Liu, Yao and Jiang, Haoming and Li, Lihong and Yun, Hyokun and Zhao, Tuo},
  booktitle = {Advances in Neural Information Processing Systems},
  volume    = {38},
  year      = {2025}
}

@article{ding2026adarubric,
  title   = {{AdaRubric}: Task-Adaptive Rubrics for Reliable {LLM} Agent Evaluation and Reward Learning},
  author  = {Ding, Liang},
  journal = {arXiv preprint arXiv:2603.21362},
  year    = {2026}
}

@inproceedings{liu2026skyworkrewardv2,
  title     = {{Skywork-Reward-V2}: Scaling Preference Data Curation via {Human--AI} Synergy},
  author    = {Liu, Chris Yuhao and Zeng, Liang and Xiao, Yuzhen and He, Jujie and Liu, Jiacai and Wang, Chaojie and Yan, Rui and Shen, Wei and Zhang, Fuxiang and Xu, Jiacheng and Liu, Yang},
  booktitle = {International Conference on Learning Representations},
  year      = {2026}
}

@inproceedings{badshah2026scope,
  title     = {{SCOPE}: Selective Conformal Optimized Pairwise {LLM} Judging},
  author    = {Badshah, Sher and Emami, Ali and Sajjad, Hassan},
  booktitle = {International Conference on Machine Learning},
  year      = {2026}
}

@article{sahoo2025quantitative,
  title   = {Quantitative {LLM} Judges},
  author  = {Sahoo, Aishwarya and Karnuthala, Jeevana Kruthi and Budhwani, Tushar Parmanand and Agarwal, Pranchal and Vaidyanathan, Sankaran and Siu, Alexa and Dernoncourt, Franck and Healey, Jennifer and Lipka, Nedim and Rossi, Ryan A. and Bhattacharya, Uttaran and Kveton, Branislav},
  journal = {arXiv preprint arXiv:2506.02945},
  year    = {2025}
}

@inproceedings{hu2022lora,
  title     = {{LoRA}: Low-Rank Adaptation of Large Language Models},
  author    = {Hu, Edward J. and Shen, Yelong and Wallis, Phillip and Allen-Zhu, Zeyuan and Li, Yuanzhi and Wang, Shean and Wang, Lu and Chen, Weizhu},
  booktitle = {International Conference on Learning Representations},
  year      = {2022}
}

@article{fedus2022switch,
  title   = {Switch Transformers: Scaling to Trillion Parameter Models with Simple and Efficient Sparsity},
  author  = {Fedus, William and Zoph, Barret and Shazeer, Noam},
  journal = {Journal of Machine Learning Research},
  volume  = {23},
  number  = {120},
  pages   = {1--39},
  year    = {2022}
}

@article{li2024mixlora,
  title   = {{MixLoRA}: Enhancing Large Language Models Fine-Tuning with {LoRA}-based Mixture of Experts},
  author  = {Li, Dengchun and Ma, Yingzi and Wang, Naizheng and Ye, Zhengmao and Cheng, Zhiyuan and Tang, Yinghao and Zhang, Yan and Duan, Lei and Zuo, Jie and Yang, Cal and Tang, Mingjie},
  journal = {arXiv preprint arXiv:2404.15159},
  year    = {2024}
}

@inproceedings{feng2024mixture,
  title     = {{Mixture-of-LoRAs}: An Efficient Multitask Tuning Method for Large Language Models},
  author    = {Feng, Wenfeng and Hao, Chuzhan and Zhang, Yuewei and Han, Yu and Wang, Hao},
  booktitle = {Proceedings of the 2024 Joint International Conference on Computational Linguistics, Language Resources and Evaluation (LREC-COLING 2024)},
  pages     = {11371--11380},
  publisher = {ELRA and ICCL},
  year      = {2024}
}

@article{zellinger2024controlled,
  title   = {Efficiently Deploying {LLM}s with Controlled Risk},
  author  = {Zellinger, Michael J. and Thomson, Matt},
  journal = {arXiv preprint arXiv:2410.02173},
  year    = {2024}
}

@inproceedings{rabanser2025gatekeeper,
  title     = {Gatekeeper: Improving Model Cascades Through Confidence Tuning},
  author    = {Rabanser, Stephan and Rauschmayr, Nathalie and Kulshrestha, Achin and Poklukar, Petra and Jitkrittum, Wittawat and Augenstein, Sean and Wang, Congchao and Tombari, Federico},
  booktitle = {Advances in Neural Information Processing Systems},
  volume    = {38},
  year      = {2025}
}

@inproceedings{melo2026epistemic,
  title     = {Epistemic Uncertainty Quantification to Improve Decisions from Black-Box Models},
  author    = {Melo, S{\'e}bastien and Varoquaux, Ga{\"e}l and Le Morvan, Marine},
  booktitle = {International Conference on Learning Representations},
  year      = {2026}
}

@misc{openassistantdeberta,
  author       = {{OpenAssistant}},
  title        = {{DeBERTa-v3} Reward-Model Checkpoints},
  year         = {2023},
  howpublished = {\url{https://huggingface.co/OpenAssistant/reward-model-deberta-v3-base}},
  note         = {Base and large public model cards}
}

@article{chow1970optimum,
  title   = {On Optimum Recognition Error and Reject Tradeoff},
  author  = {Chow, C. K.},
  journal = {IEEE Transactions on Information Theory},
  volume  = {16},
  number  = {1},
  pages   = {41--46},
  year    = {1970},
  doi     = {10.1109/TIT.1970.1054406}
}

@inproceedings{geifman2017selective,
  title     = {Selective Classification for Deep Neural Networks},
  author    = {Geifman, Yonatan and El-Yaniv, Ran},
  booktitle = {Advances in Neural Information Processing Systems},
  volume    = {30},
  year      = {2017}
}

@inproceedings{madras2018predict,
  title     = {Predict Responsibly: Improving Fairness and Accuracy by Learning to Defer},
  author    = {Madras, David and Pitassi, Toniann and Zemel, Richard},
  booktitle = {Advances in Neural Information Processing Systems},
  volume    = {31},
  year      = {2018}
}

@inproceedings{mozannar2020consistent,
  title     = {Consistent Estimators for Learning to Defer to an Expert},
  author    = {Mozannar, Hussein and Sontag, David},
  booktitle = {International Conference on Machine Learning},
  pages     = {7076--7087},
  year      = {2020}
}

@article{arora2025healthbench,
  title   = {{HealthBench}: Evaluating Large Language Models Towards Improved Human Health},
  author  = {Arora, Rahul K. and Wei, Jason and Soskin Hicks, Rebecca and Bowman, Preston and Qui{\~n}onero-Candela, Joaquin and Tsimpourlas, Foivos and Sharman, Michael and Shah, Meghan and Vallone, Andrea and Beutel, Alex and Heidecke, Johannes and Singhal, Karan},
  journal = {arXiv preprint arXiv:2505.08775},
  year    = {2025}
}

@article{lambert2024tulu3,
  title   = {{T\"ulu} 3: Pushing Frontiers in Open Language Model Post-Training},
  author  = {Lambert, Nathan and Morrison, Jacob and Pyatkin, Valentina and Huang, Shengyi and Ivison, Hamish and Brahman, Faeze and Miranda, Lester James V. and Liu, Alisa and Dziri, Nouha and Lyu, Shane and Gu, Yuling and Malik, Saumya and Graf, Victoria and Hwang, Jena D. and Yang, Jiangjiang and Le Bras, Ronan and Tafjord, Oyvind and Wilhelm, Chris and Soldaini, Luca and Smith, Noah A. and Wang, Yizhong and Dasigi, Pradeep and Hajishirzi, Hannaneh},
  journal = {arXiv preprint arXiv:2411.15124},
  year    = {2024}
}
\bibliographystyle{plainnat}

\section*{Ethics Statement}

This work uses public model outputs and benchmark annotations, including
physician labels released with HealthBench; it collects no new human-subject
data. Automated evaluators can reduce review cost, but they can also encode
contested policies, centralize judgment, or invite more confidence than a
benchmark audit warrants. Criteria should remain versioned and attributable,
and affected users should have a route of appeal. Coverage measured on these
benchmarks is not a reason to remove human review from legal, medical,
employment, or other high-impact decisions.

\appendix
\section{Reproducibility Details}

All splits use seed 20260728. Direct-assessment adapters train with batch
size 4, gradient accumulation 8, and 1,024-token inputs unless marked
otherwise. Table~\ref{tab:training-recipes} records initialization, learning
rate, and optimizer updates for every specialization control. The original
experiments ran on two RTX 5090 GPUs (32,607 MiB each) with PyTorch 2.8.0,
CUDA 12.8, and NCCL 2.27.3. The revision experiments (the data-budget
grid, the HealthBench study, and cross-family scoring) ran on one RTX 4090
(24,564 MiB) with PyTorch 2.11.0 and CUDA 12.8, with recipes and seeds
unchanged. Prediction-level outputs, split manifests, training metadata,
threshold audits, paired-bootstrap results, and all 20-seed outputs for each
cascade configuration are included in the artifact. Dataset file sizes and
SHA-256 hashes are recorded in the download manifest.

\begin{table}[h]
\caption{Training recipes for the direct-assessment specialization controls.
``Shared updates'' are full-corpus optimizer steps completed before any
family split; ``family updates'' are summed across routed experts.}
\label{tab:training-recipes}
\centering
\small
\begin{tabular}{llrrr}
\toprule
Design & Initialization & LR & Shared updates & Family updates\\
\midrule
Monolith r8 & base & $2e{-4}$ & 1,489 & --\\
$K=8$, scratch & base & $2e{-4}$ & 0 & 1,494\\
$K=8$, scratch, matched LR & base & $5e{-5}$ & 0 & 1,494\\
$K=8$, 500 each & base & $2e{-4}$ & 0 & 4,000\\
$K=8$, shared init. & monolith r8 & $5e{-5}$ & 1,489 & 1,494\\
Monolith r64 & base & $2e{-4}$ & 1,489 & --\\
\bottomrule
\end{tabular}
\end{table}

\begin{table}[h]
\caption{Criterion-family granularity and optimization controls. Adapter
counts are millions; ``Shift'' is the unseen-criterion split and
``External'' is Feedback-Bench. Coverage is the locked-test acceptance rate
under the calibration-selected Bonferroni policy at the 5\% risk bound;
$^\dagger$ marks a policy whose one-sided upper bound exceeds 5\%.}
\label{tab:granularity}
\centering
\small
\begin{tabular}{lrrrrrr}
\toprule
Design & Stored & Active & Test & Shift & External & Coverage\\
\midrule
Monolith r8 & 5.05 & 5.05 & 76.38 & 77.85 & 77.60 & 24.44\\
$K=4$, scratch & 20.21 & 5.05 & 71.81 & 73.78 & 74.90 & 17.79\\
$K=8$, scratch & 40.41 & 5.05 & 66.34 & 67.72 & 69.10 & 5.43\\
$K=8$, scratch, lr $5e{-5}$ & 40.41 & 5.05 & 56.91 & 58.47 & 58.00 & 0.00\\
$K=8$, 500 each & 40.41 & 5.05 & 72.56 & 73.74 & 79.40 & 11.76$^\dagger$\\
$K=8$, shared init. & 40.41 & 5.05 & 76.85 & 78.14 & 78.90 & 24.90\\
Monolith r64 & 40.38 & 40.38 & \textbf{77.50} & \textbf{78.97} &
\textbf{82.10} & \textbf{30.14}\\
\bottomrule
\end{tabular}
\end{table}

\begin{table}[h]
\caption{Data-budget grid. Top: locked-test accuracy for the five designs;
``$+$epoch'' is the continued-pooled control (one further pooled epoch at
$5\times10^{-5}$ from the same budget's monolith). Bottom: locked-test
acceptance under the Bonferroni policy at the 5\% bound; $^\dagger$ marks a
policy whose one-sided upper bound exceeds 5\% on the locked test.
Full-data entries repeat Table~\ref{tab:granularity}.}
\label{tab:scaling}
\centering
\small
\begin{tabular}{rrrrrrr}
\toprule
Budget & Ex./family & Monolith & $+$epoch & $K{=}4$ scratch
& $K{=}8$ scratch & $K{=}8$ shared\\
\midrule
1/16 & 372 & 53.30 & 57.74 & 38.79 & 26.49 & 53.83\\
1/8 & 744 & 65.91 & 67.19 & 48.12 & 31.79 & 65.58\\
1/4 & 1,489 & 69.44 & 70.95 & 57.43 & 44.51 & 70.03\\
1/2 & 2,977 & 73.91 & 74.31 & 65.84 & 57.31 & \textbf{74.77}\\
1 & 5,955 & 76.38 & -- & 71.81 & 66.34 & \textbf{76.85}\\
\bottomrule
\end{tabular}

\vspace{0.6em}

\begin{tabular}{rrrrr}
\toprule
Budget & Ex./family & Monolith cov. & $K{=}8$ scratch cov.
& $K{=}8$ shared cov.\\
\midrule
1/16--1/8 & 372--744 & 0.0 & 0.0 & 0.0\\
1/4 & 1,489 & 1.9 & 0.0 & 1.5$^\dagger$\\
1/2 & 2,977 & 15.0 & 0.0 & 15.0\\
1 & 5,955 & 24.4 & 5.4 & \textbf{24.9}\\
\bottomrule
\end{tabular}
\end{table}

Budget labels are nominal means (fraction $\times$ 5,954.8). Hash-based
sampling yields actual training means of 374, 741, 1,492, and 2,960
examples per family, and family sizes vary widely around them: 188--992 at
the 1/16 budget and 1,550--7,739 at 1/2. The continued-pooled control
starts from each budget's monolith and runs one further pooled epoch at
$5\times10^{-5}$, matching the extra optimization the warm-started bank
receives. The $K{=}4$ banks use a four-cluster router fitted with the same
procedure and seed on the same sampled rows. Every training run in the
sweep uses the single seed 20260728, so the criterion-clustered intervals
reflect sampling variability, not run-to-run training variability.

\begin{table}[h]
\caption{HealthBench paired differences, conversation-clustered bootstrap
(10,000 resamples). Positive favors the second-listed condition.}
\label{tab:healthbench-boot}
\centering
\small
\begin{tabular}{llrrr}
\toprule
Split & Comparison & Diff. (pp) & 95\% interval & $p$\\
\midrule
Test & ablated $\to$ rubric & $+1.60$ & $[+0.46,\,+2.74]$ & 0.004\\
Test & rubric $\to$ wrong rubric & $-4.31$ & $[-5.57,\,-3.05]$ & $2\times10^{-4}$\\
Unseen & ablated $\to$ rubric & $+0.08$ & $[-0.34,\,+0.51]$ & 0.70\\
Unseen & rubric $\to$ wrong rubric & $-2.28$ & $[-2.80,\,-1.77]$ & $2\times10^{-4}$\\
\bottomrule
\end{tabular}
\end{table}

\section{Additional Limitations}
\label{sec:add-limitations}

The compiler uses English lexical features. Dense semantic routing,
multilingual criteria, joint multi-expert training, and other
parameter-efficient modules may shift the specialization tradeoff. The
ordinal head produces no written critique.

Risk results are retrospective audits on frozen splits. Wilson bounds do not
establish prospective guarantees under distribution shift, and the threshold
procedure scans candidate cutoffs. High-impact applications need
pre-specified risk control, monitoring, and mandatory human gates. The 20
RewardBench repartitions reuse the same tasks; they measure sensitivity to
partition assignment, not sampling uncertainty across independent datasets.
We therefore use exact binomial bounds within each held-out partition and do
not treat across-seed quantiles as population confidence intervals.
Normalized parameter compute is not a measurement of energy or monetary
cost, and the workflow calculation is not evidence of achieved
organizational throughput.

\section{Risk--Coverage Curves}

\begin{figure}[H]
\centering
\includegraphics[width=0.62\linewidth]{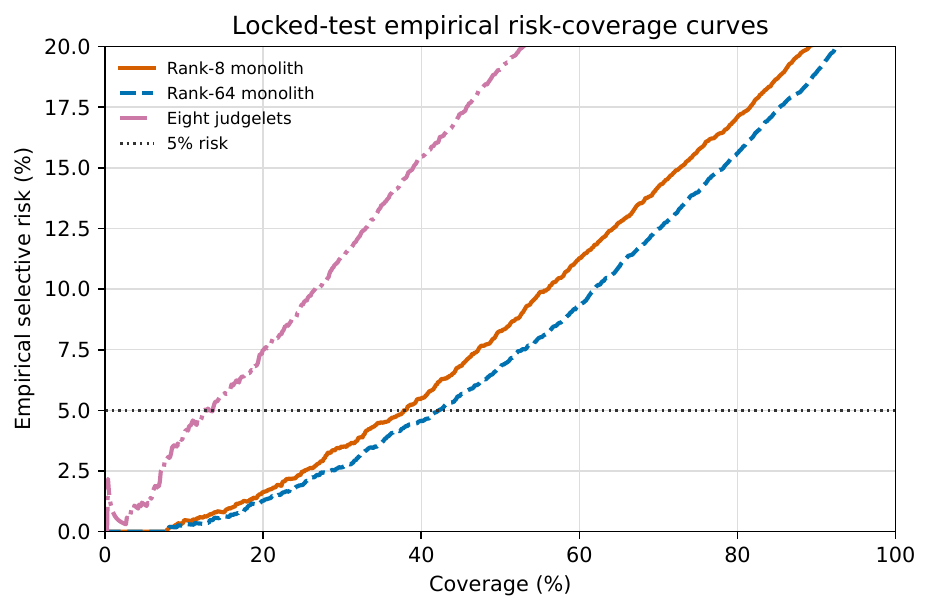}
\caption{Empirical locked-test risk as coverage increases. The fragmented
eight-judgelet bank loses its low-risk region much earlier than either
monolith; line styles preserve the comparison without color. Thresholded
coverage at the 5\% bound is reported in Table~\ref{tab:granularity}.}
\end{figure}

\section{Workflow Sensitivity}

Selective coverage is the fraction of decisions that can leave a review
queue. For interpretation only, we substitute measured coverage and
accepted-set risk into an Amdahl-style sensitivity model
\citep{amdahl1967validity}.\footnote{Amdahl's law motivates the
serial-fraction term; $e$ and $crk$ are diagnostic extensions introduced
here.}
\[
S=\{(1-c)+c/s+e+crk\}^{-1},
\]
where $s$ is automatic-review speedup, $e$ evaluator overhead, and $k$
escaped-error rework. At $s=20$, $e=0.02$, and $k=1$, moving from the
rubric-ablated Bonferroni policy to the rank-8 and rank-64 rubric-aware
policies raises implied speedup from 1.171$\times$ to 1.258$\times$ and
1.345$\times$. This is a sensitivity calculation, not a field measurement;
the artifact provides the full grid.

\end{document}